%% file: main.tex
\input{constants}

\arxiv

\documentclass[11pt]{article}

\input{acl_header}

\title{\paperTitle}
\author{\authorBlock}

\begin{document}
    \maketitle

{
  \renewcommand{\thefootnote}%
  {\fnsymbol{footnote}}
  \footnotetext[1]{Equal contribution.}
  \footnotetext[2]{Corresponding Author.}
}

\begin{abstract}
Named-entity recognition (NER) detects texts with predefined semantic labels and is an essential building block for natural language processing (NLP).
Notably, recent NER research focuses on utilizing massive extra data, including pre-training corpora and incorporating search engines.
However, these methods suffer from high costs associated with data collection and pre-training, and additional training process of the retrieved data from search engines.
To address the above challenges, we completely frame NER as a machine reading comprehension (MRC) problem, called NER-to-MRC, by leveraging MRC with its ability to exploit existing data efficiently. 
Several prior works have been dedicated to employing MRC-based solutions for tackling the NER problem, several challenges persist: i) the reliance on manually designed prompts; ii) the limited MRC approaches to data reconstruction, which fails to achieve performance on par with methods utilizing extensive additional data. 
Thus, our NER-to-MRC conversion consists of two components: i) transform the NER task into a form suitable for the model to solve with MRC in a efficient manner; ii) apply the MRC reasoning strategy to the model. 
We experiment on $6$ benchmark datasets from three domains and achieve state-of-the-art performance without external data, up to $11.24\%$ improvement on the WNUT-$16$ dataset.

\end{abstract}

\input{sections/0introduction}

\input{sections/1related_work}

\input{sections/2method}

\input{sections/3experiment}

\input{sections/4conclusion}

\input{sections/5limitations}

\input{sections/6ethical}

{\small
\bibliography{anthology,custom}
\bibliographystyle{acl_natbib}
}

\clearpage
\input{sections/appendix}

\end{document}

%% file: constants.tex
\def\paperTitle{
NER-to-MRC: Named-Entity Recognition Completely Solving as Machine Reading Comprehension
}

\def\authorBlock{
    Yuxiang Zhang$^\textnormal{1}$\footnotemark[1] \qquad
    Junjie Wang$^\textnormal{1}$\footnotemark[1] \qquad
    Xinyu Zhu$^\textnormal{2}$ \\
    \textbf{
    Tetsuya Sakai$^\textnormal{1}$ \qquad
    Hayato Yamana$^\textnormal{1}$\footnotemark[2]
    } \\
    $^\textnormal{1}$Waseda University \quad
    $^\textnormal{2}$Tsinghua University \quad
    \\
    {\tt\small joel0495@asagi.waseda.jp} \quad 
    {\tt\small wjj1020181822@toki.waseda.jp} \\
    {\tt\small zhuxy21@mails.tsinghua.edu.cn} \quad
    {\tt\small tetsuyasakai@acm.org} \quad
    {\tt\small yamana@yama.info.waseda.ac.jp}
}

\newif\ifreview 
\newif\ifarxiv \newcommand{\arxiv}{\arxivtrue}
\newif\ifcamera 

%% file: acl_header.tex
\ifreview \usepackage[review]{acl} \fi
\ifarxiv \usepackage{acl} \fi
\ifcamera \usepackage[final]{acl} \fi

\usepackage{times}
\usepackage{latexsym}
\usepackage[T1]{fontenc}

\usepackage[utf8]{inputenc}

\usepackage{microtype}

\usepackage{graphicx}
\usepackage{amsmath}
\usepackage{amssymb}
\usepackage{booktabs}
\usepackage{amsfonts}
\usepackage{algorithm}
\usepackage{algorithmicx}
\usepackage{algpseudocode}
\usepackage{footnote}

\usepackage{xcolor}
\usepackage{comment}
\usepackage{amsthm}
\usepackage{bbm} 
\usepackage{subcaption}
\usepackage{multirow}
\usepackage{comment}
\usepackage{makecell} 
\usepackage{tabularx}
\usepackage[normalem]{ulem}
\useunder{\uline}{\ul}{}
\usepackage{booktabs}
\usepackage[switch]{lineno}

\usepackage{url}
\usepackage{xspace}

\usepackage[export]{adjustbox}
\usepackage{paralist}

\newcommand{\nbf}[1]{{\noindent \textbf{#1}}}

\newcommand{\eat}[1]{}

\newcommand{\cbit}{\begin{compactitem}}
\newcommand{\ceit}{\end{compactitem}}
\newcommand{\cben}{\begin{compactenum}}
\newcommand{\ceen}{\end{compactenum}}

\DeclareMathOperator*{\argmax}{argmax}

\usepackage[capitalize]{cleveref}
\crefname{section}{Sec.}{Secs.}
\crefname{table}{Table}{Tables}
\crefname{figure}{Fig.}{Figs.}
\crefname{algocf}{alg.}{algs.}
\Crefname{algocf}{Algorithm}{Algorithms}

%% file: sections/0introduction.tex
\section{Introduction}
\label{sec:introduction}

\begin{figure}[!tp]
    \centering
    \includegraphics[width=0.49\textwidth]{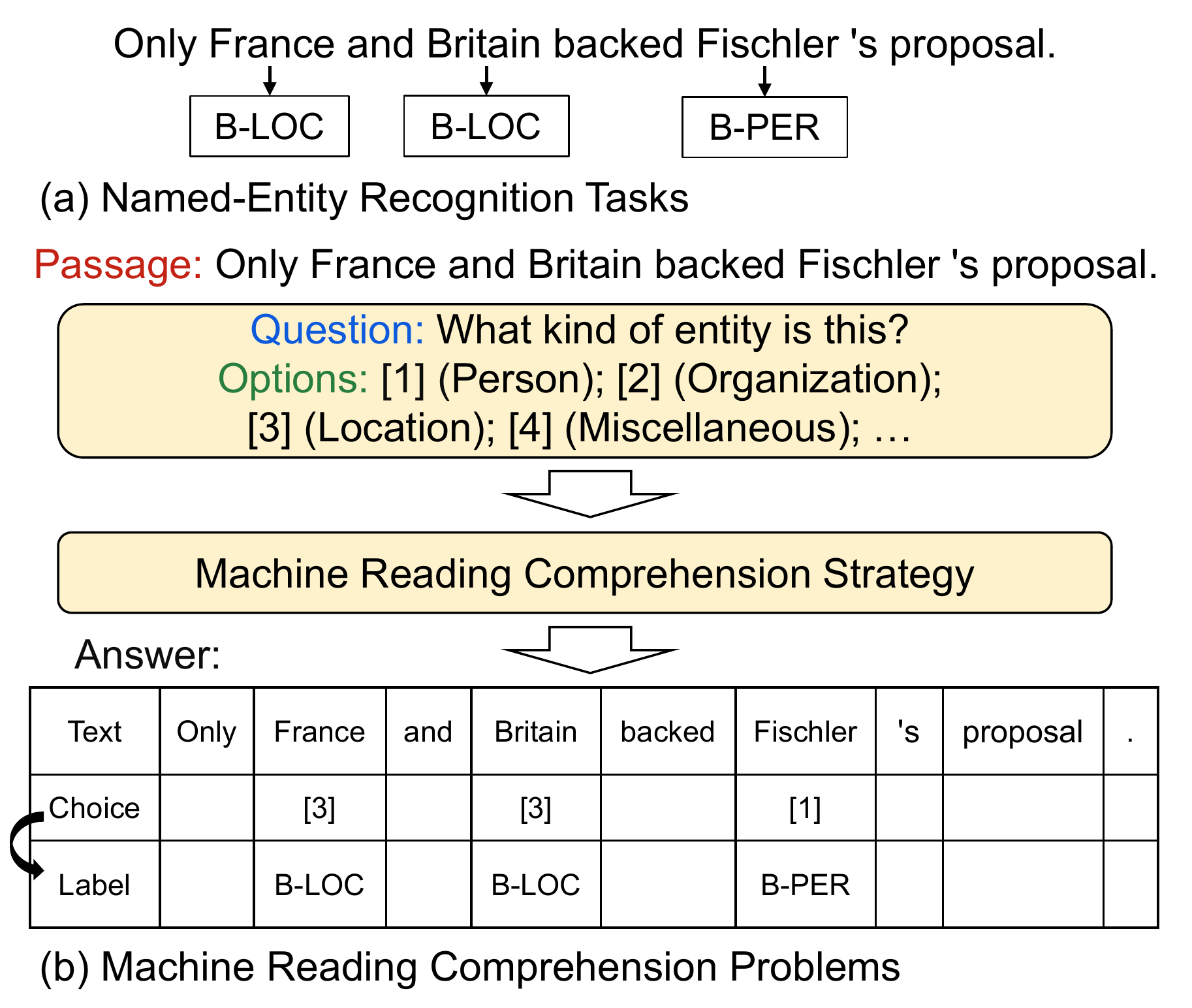}
    \caption{Comparison of the name-entity recognition task and machine reading comprehension problem, where the example text and related labels are the same as CoNLL-2003 dataset\cite{DBLP:conf/conll/SangM03@conll2003}. For presentation purposes, we omit the entity label ``Other''. (a) shows a typical solution for NER tasks. (b) presents our NER-to-MRC framework.}
    \label{fig:introduction}
\end{figure}

A fundamental topic in natural language processing (NLP) is named entity recognition (NER), which aims to detect the text with predefined semantic labels, such as a person, a position, and others~\cite{DBLP:journals/tkde/LiSHL22@ner-survey}.
Given its importance, many methods have been published for this task and most of them can be viewed within a sequence labeling framework
~\cite{DBLP:conf/acl/LiFMHWL20@bertmrc,DBLP:conf/acl/WangJBWHHT20@clkl,DBLP:conf/acl/FuHL20@spanner,DBLP:conf/naacl/DevlinCLT19@bert} that tags the input words, as shown in~\cref{fig:introduction} (a).
For example, with a given sentence, a properly working NER system should recognize three named entities with its inside–outside–beginning (IOB) format tagging~\cite{DBLP:conf/acl-vlc/RamshawM95@IOB}: ``France'' is a location entity (B-LOC), ``Britain'' is a location entity (B-LOC), and ``Fischler'' is a person entity (B-PER).
NER tasks provide only a single sequence with limited information as the model's input. Consequently, several approaches~\cite{DBLP:conf/acl/WangJBWHHT20@clkl,DBLP:journals/jucs/JM15@retrieval-1,DBLP:journals/corr/abs-2210-07523@retrieval-2,DBLP:conf/emnlp/WangCJXTL22@retrieval-3} consider how to efficiently utilize additional information to enhance the understanding of the input sequence, thereby improving model performance. However, the effective retrieval of supplementary information and training on this extra data require additional cost and effort.

The above contradiction poses a challenge, and hence the focus of our paper is to address this.  
The ability of reading comprehension can reflect human understanding and reasoning skills~\cite{Davis_1944@RC, Trassi_Oliveira_Inácio_2019@RC-cloze}, such as question-answering or multiple-choice. 
Additionally, recent studies show that machine reading comprehension (MRC) can improve various tasks, such as natural language inference and text classification~\cite{DBLP:journals/corr/abs-2210-08590@unimc,DBLP:conf/iclr/WeiBZGYLDDL22@flan,DBLP:conf/iclr/SanhWRBSACSRDBX22@t0}.
Inspired by these, we study the problem of how to harness MRC to address NER, resulting in bridging MRC reasoning strategy with NER task.
To begin with, we require high-quality data reconstruction to convert NER as an MRC problem. 
Our preferred method for addressing this construction involves the use of artificially designed NER-to-MRC queries, including BERT-MRC~\cite{DBLP:conf/acl/LiFMHWL20@bertmrc} and KGQA~\cite{DBLP:journals/health/BanerjeePDB21@KGQA}. However, this solution faces the following challenges,
i) low-transferability: the lack of a unified data reconstruction paradigm makes it difficult to migrate to new datasets.
ii) hand-crafted: high reproduction difficulty and unstable performance due to hand-crafted question templates; 
iii) insufficient information: the input ignores label semantics.

To address the above challenges, we frame the data construction
as a multiple choice (MC) problem.
Before detailing the solution, we first present an example of MC format as follows:  given a $(passage, question, options)$ triplet, the system chooses ``Location'' as the entity label for ``France''.
The MC format overcomes the counter-intuitive challenge mentioned above by allowing the model to predict possible entity labels through text descriptions.
In essence, what matters the most for building the MC format is to generate appropriate questions for prompting language models.
Inspired by the design principle (a less manual process) of UniMC~\cite{DBLP:journals/corr/abs-2210-08590@unimc}, we only provide a single prompt question, which implies stable results as opposed to inconsistent question templates in existing schemes to resolve the hand-crafted challenge. 
On the other hand, to address the insufficient information challenge, our design introduces label information as options, which conveys essential semantics in a variety of low-resource scenarios~\cite{luo-etal-2021-dont@dont-miss-label,DBLP:conf/acl/MuellerKRMMZR22@label-semantic}.

Recent state-of-the-art (SoTA) NER works rely on the assistance of external data, such as retrieved text from the Google search engine~\cite{DBLP:conf/acl/WangJBWHHT20@clkl} and pre-training data~\cite{DBLP:conf/emnlp/YamadaASTM20@luke}.
To achieve performance comparable to state-of-the-art external context retrieving approaches without relying on additional retrieved data, merely considering MRC methods in the data reconstruction phase is insufficient. We believe that MRC techniques can learn missing information from the dataset without any extra data. Therefore, we further integrated MRC reasoning strategies into the NER tasks.
Specifically, we introduce a powerful human reading strategy, HRCA~\cite{DBLP:conf/lrec/ZhangY22@hrca} (details in~\cref{sec:network}), instead of only inserting a Pre-trained Language Model (PLM) as BERT-MRC. 
To this end, we transfer the model choice to entity labels by matching the options.

To evaluate our framework, we carry out numerous experiments on $6$ challenging NER benchmarks, including three domains with general and specialized vocabulary.
The results demonstrate that our approach improves SoTA baselines, such as WNUT-$16$~\cite{DBLP:conf/aclnut/StraussTRMX16/@wnut16} ($+ 11.24\%$), WNUT-$17$~\cite{DBLP:conf/aclnut/DerczynskiNEL17@wnut17} ($+ 0.76\%$), BC$5$CDR~\cite{DBLP:journals/biodb/LiSJSWLDMWL16@bc5cdr} ($+ 1.48\%$), NCBI~\cite{DBLP:journals/jbi/DoganLL14@ncbi} ($+ 1.09\%$), CoNLL-$2003$~\cite{DBLP:conf/conll/SangM03@conll2003} ($+ 0.33\%$) and CoNLL$++$~\cite{DBLP:conf/emnlp/WangSLLLH19@conll++} ($+ 0.44\%$).
Note that our MRC method achieves such performance without any extra data, which suggests the potential of mining text for intrinsic connections to complete complex tasks.

In summary, our contributions are:
\begin{itemize}
    \item We propose a new NER-to-MRC reconstruction solution by introducing a simple yet stable data reconstruction rule.
    \item We apply MRC strategies to solve NER problems without retrieval systems or pre-training NER data, resulting in a concise process.
    \item Our approach shows the SoTA performance on multiple popular NER datasets, including three domains with general (WNUT-$16$, WNUT-$17$, CoNLL-$2003$ and CoNLL$++$) and specialized (BC$5$CDR and NCBI) vocabulary. 
\end{itemize}

%% file: sections/1related_work.tex
\section{Related Work}
\label{sec:related_work}

We introduce the general development trend of the NER field in~\cref{sec:ner}. Treating NLP tasks other than MRC as MRC problems can enhance the performance of corresponding tasks. 
We demonstrate how such schemes apply the MRC paradigm to other NLP tasks in~\cref{sec:mrc}.

\subsection{Named Entity Recognition (NER)}
\label{sec:ner}

NER is designed to detect words from passages by predefined entity labels~\cite{DBLP:journals/tkde/LiSHL22@ner-survey}, which serves as the foundation of complicated applications such as machine translation.
The BiLSTM architecture is the most commonly used architecture for solving NER tasks in the early days of deep learning~\cite{DBLP:journals/tkde/LiSHL22@ner-survey}. In the bi-directional LSTM, each word's representation can be derived from the contextual representation that connects its left and right directions, which is advantageous for many tagging applications~\cite{DBLP:conf/naacl/LampleBSKD16@bilstm_crf}.
Later, with the development of pre-trained language models represented by BERT~\cite{DBLP:conf/naacl/DevlinCLT19@bert}, transformer-based pre-trained models quickly became the preferred model for the NLP area~\cite{DBLP:journals/corr/abs-2106-04554@transformer}. The embeddings learned by these transformer-based pre-trained language models are contextual and trained on a large corpus. As a result, for NER tasks that value input representation, pre-trained language models rapidly become a new paradigm.
In recent years, introducing external data to PLMs becomes dominant and shows powerful contextual representations, such as NER-BERT~\cite{DBLP:journals/corr/abs-2112-00405@ner-bert}, LUKE~\cite{DBLP:conf/emnlp/YamadaASTM20@luke} and CL-KL~\cite{DBLP:conf/acl/WangJBWHHT20@clkl}. 
LUKE~\cite{DBLP:conf/emnlp/YamadaASTM20@luke} proposes an entity-aware self-attention mechanism and a pre-training task to predict masked words and entities in a sizeable entity-annotated corpus.
CL-KL~\cite{DBLP:conf/acl/WangJBWHHT20@clkl} finds the external contexts of query sentences by employing a search engine and then processes them by a cooperative learning method.
However, large-scale external datasets consume considerable collection time and even labor costs.
Therefore, we explore an extra-data-free framework by MRC strategies after transferring NER tasks to MRC problems.

\subsection{Enhancing NLP Tasks via MRC perspective}
\label{sec:mrc}

Treating NLP tasks other than MRC as MRC problems strengthens neural networks' reasoning processes, including event extraction~\cite{DBLP:conf/emnlp/DuC20@ee-qa,DBLP:conf/emnlp/LiuCLBL20@ee-mrc}, relation extraction~\cite{DBLP:conf/acl/LiYSLYCZL19@er-qa,DBLP:conf/conll/LevySCZ17@zero-er-mrc}, and named-entity recognition~\cite{DBLP:conf/acl/LiFMHWL20@bertmrc}.
The current mainstream approaches~\cite{DBLP:conf/acl/LiFMHWL20@bertmrc,DBLP:journals/health/BanerjeePDB21@KGQA,DBLP:journals/jbi/SunYWZLW21@BioMRC,DBLP:conf/emnlp/DuC20@ee-qa,DBLP:conf/acl/LiYSLYCZL19@er-qa,DBLP:conf/conll/LevySCZ17@zero-er-mrc,DBLP:conf/emnlp/LiuCLBL20@ee-mrc,DBLP:conf/emnlp/XueYZLZW20@CoFEE,DBLP:conf/naacl/ShrimalJMY22@NERMQMRC} utilizes data reconstruction to address NER tasks using MRC methodology. Specifically, those approach involves restructuring the input data into MRC format by incorporating MRC-style question prompts. 
These prompts encompass the direct utilization of entities as questions, human-designed question templates and unsupervised generated questions.
However, those methods have missed important label semantics, which describe the entity labels from annotation guidelines in each NER dataset.
On the other hand, even though all of these approaches explore the utilization of MRC for solving the NER problem, they merely employ the MRC scheme during the data reconstruction phase, rather than effectively leveraging the most critical reasoning strategy of MRC for the NER task.
To address the problems above, our method inferences the possible entity type through: 
i) introducing label information; 
ii) applying MRC reasoning strategies.

\begin{figure*}[!tp]
\centering
\includegraphics[width=0.98\textwidth]{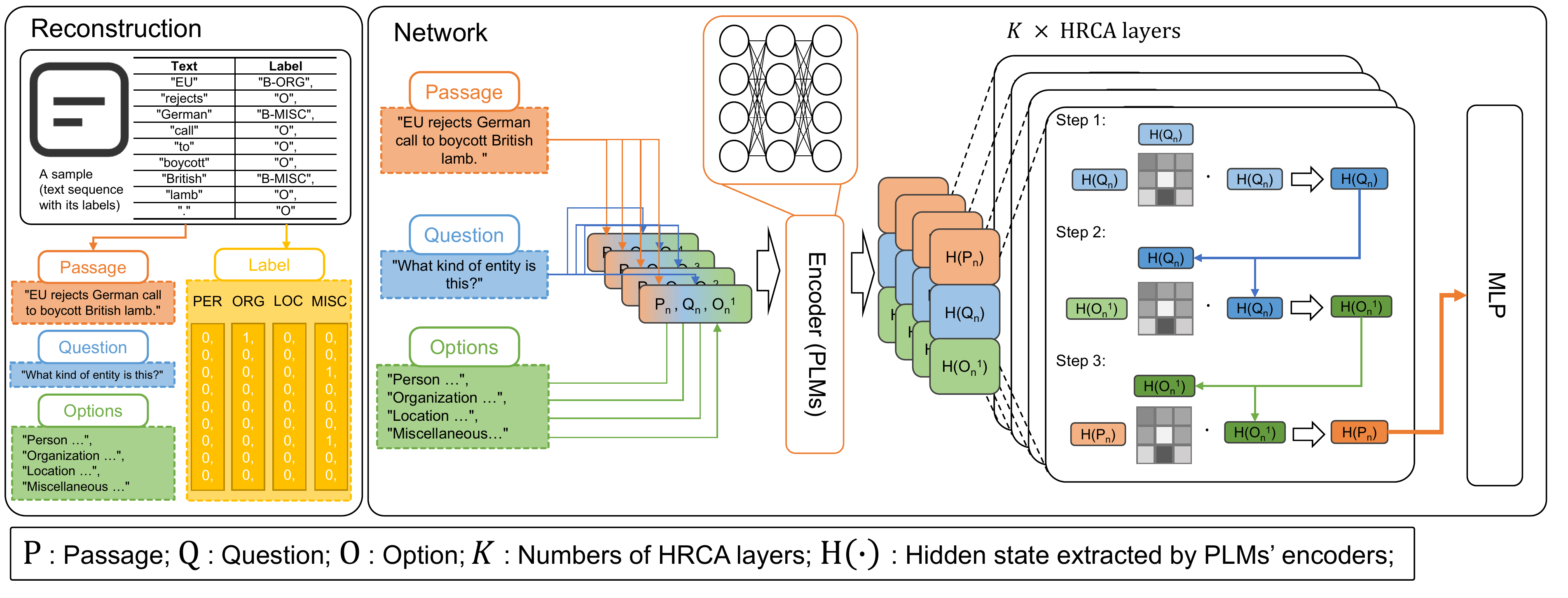}
\caption{
The proposed NER-to-MRC framework. A data sample from the NER tasks will first go through step (a) reconstruction, and then into step (b) network for learning a powerful representation.
}
\label{fig:model_framework}
\end{figure*}

%% file: sections/2method.tex
\section{Complete NER-to-MRC Conversion}
\label{sec:method}

The current field of NER faces the following challenges: i) hand-crafted design and inadequate information; ii) failure to integrate crucial MRC reasoning strategies throughout the entire inference process.
In this section, we outline the proposed NER-to-MRC framework, including input reconstruction (\cref{sec:reconstruction}) to address the hand-crafted design challenge, MRC reasoning network (\cref{sec:network}) to introduce essential MRC processes into NER tasks. We demonstrate the fine-tuning of our framework in~\cref{sec:finetuning}.

\subsection{Input Reconstruction}
\label{sec:reconstruction}

Input reconstruction is a key topic for facilitating the solution of NER tasks from a MRC perspective. We introduce a simple and instructional reconstruction rule (Details in~\cref{append:example_input_rec}).
Alternative to the question-and-answer format in previous work~\cite{DBLP:conf/acl/LiFMHWL20@bertmrc}, we consider a multiple-choice (MC) format that incorporates label information, as shown in~\cref{fig:model_framework} (a).
Given a common NER dataset, a sample includes a text sequence $X = \left\{x_{1}, x_{2}, ..., x_{n}\right\}$ with $n$ words and the corresponded entity label $Y = \left\{y_{1}, y_{2}, ..., y_{n}\right\}$ for each word.
Then, the transformed MC format ($(passage, question, options)$ triplets) is constructed by:

\nbf{Passage}: We obtain the passage part effortlessly, which comes from the text sequence $X$ in the original dataset.

\nbf{Question}: For less human-crafted processing, we only provide a universal question (``What kind of entity is this?'') for all types of entities. It is noteworthy that, due to the fixed inputs, our model produces stable results that are reproducible, in contrast to hand-crafted questions that can vary from individual to individual.

\nbf{Options}: We treat an entity type as significant input with semantic information rather than just a label. Specifically, we borrow the description of each entity label from the annotation guidelines. 

Finally, an input sample consists of a passage $P = \left\{p_{1}, p_{2}, \dots, p_{k}\right\}$, a question $Q = \left\{q_{1}, q_{2}, \dots, q_{m}\right\}$, and $N_O$ options $O = \left\{o^{i}_{1}, o^{i}_{2}, \dots, o^{i}_{n}\right\}^{N_O}_{i=1}$ with an entity description, where $k,m,n$ are the corresponding sequence lengths for passage, question, and options.
Regarding labels, we remove the ``B-'' and ``I-'' tagging and only keep the entity type itself. Then, based on the sequence length of the passage and the number of entity types, the label is processed as a binary matrix $ M_{label} \in \mathbb{B}^{k \times N_O} $ with assigning $1$ to each corresponding entity type on the labels, $0$ to the other.

\subsection{MRC Reasoning Network}
\label{sec:network}

Our framework employs PLMs with MRC strategies as the MRC reasoning network.
Remarkable advances~\cite{banditvilai2020effectiveness@mrc-strategy-effective,baier2005reading@mrc-strategy-experiment} suggest that reading methods improve participants to learn a deep understanding of reading comprehension. 
Note that the success of HRCA~\cite{DBLP:conf/lrec/ZhangY22@hrca} on MRC tasks presents a human-like reasoning flow as a MRC strategy.
However, HRCA only focuses on MRC tasks and does not support other NLP tasks.
Therefore, we apply it to our NER-to-MRC framework to enhance reasoning ability.
In particular, we firstly generate $N_O$ MRC triplets input by contacting each option and $P$ and $Q$ ($P \oplus Q \oplus O$).
Then, encoded by PLMs ($Encoder$), the hidden state $H$ is obtained as: $H(P \oplus Q \oplus O) = Encoder(P \oplus Q \oplus O)$.
After that, the $H(P \oplus Q \oplus O)$ are separated into three parts $H(P), H(Q), H(O)$ corresponding to the MRC triplet and fed into HRCA layers. We process those three hidden states in the following $3$ steps:
i) ``reviewing'' $H(Q)$ (question) by applying self-attention mechanism. 
ii) ``reading'' $H(O)$ (options) with $H(Q)$ (question) by computing cross attention weights.
iii) ``finding'' final results in $H(P)$ (passages) with $H(O)$ (options).
Additionally, we apply multi-head operation~\cite{DBLP:conf/lrec/ZhangY22@hrca} in all attention computation in the HRCA layer.
After that, rich semantics from options and questions are embedded into the hidden states of the passage part.

\begin{figure}[!tp]
    \centering
    \includegraphics[width=0.49\textwidth]{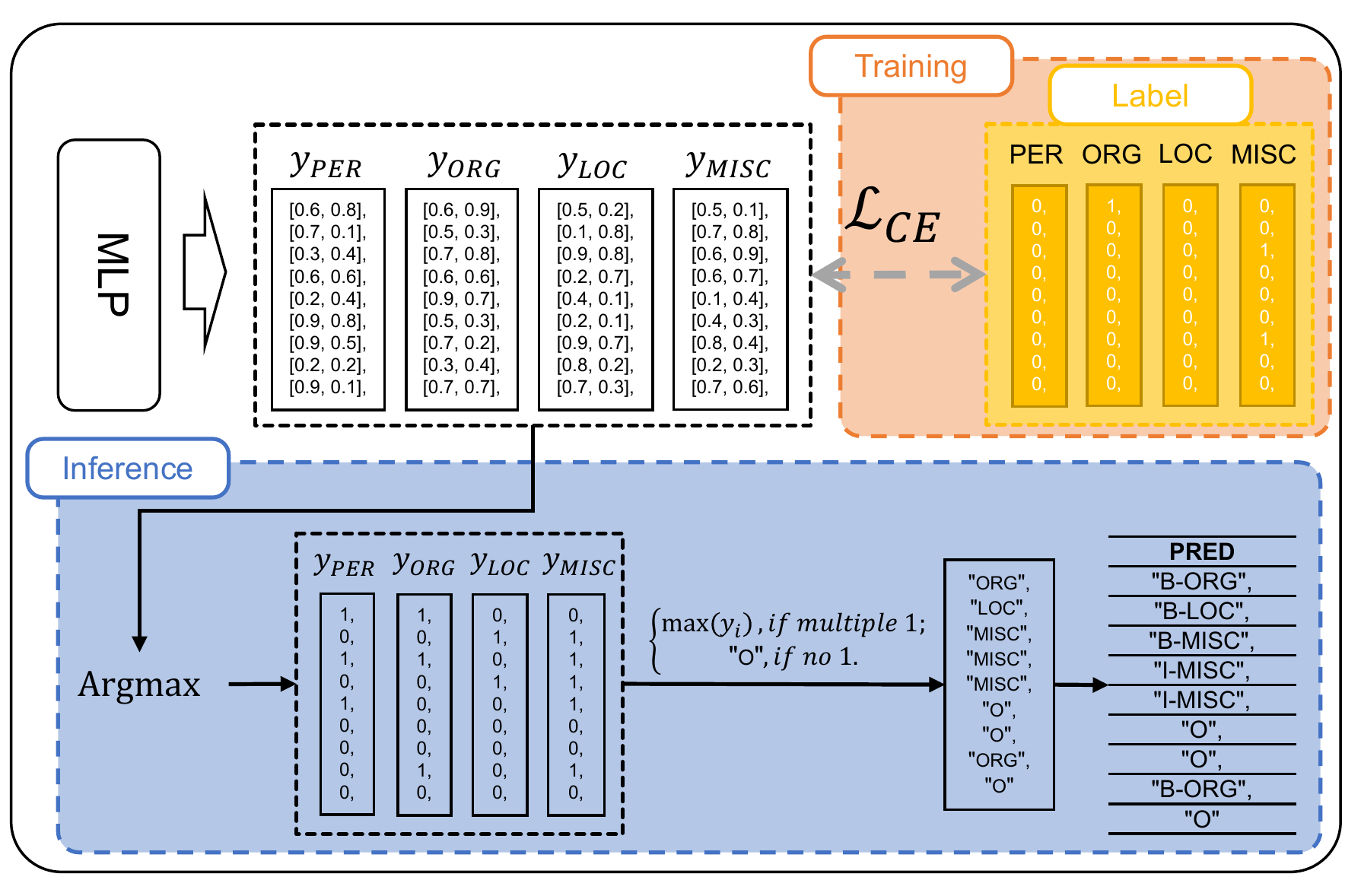}
    \caption{
    Overall fine-tuning procedure for NER-to-MRC. After training with golden data to tune all parameters, we infer it in unseen test set.
    }
    \label{fig:finetuning} 
\end{figure} 

\subsection{Fine-tuning}
\label{sec:finetuning}

For model multiple downstream datasets, we fine-tune the NER-to-MRC framework, as shown in~\cref{fig:finetuning}.
In detail, we pass the encoded hidden states of the passage to a MLP layer, which includes $N_O$ sub MLP layers.
Those sub MLP layers reduce the original dimensions to $2$, which indicates select it or not.
Therefore, after the MLP layer, we can obtain a prediction matrix $M_{pred} \in \mathbb{R}^{k \times N_O \times 2}$.

\subsubsection{Training}
\label{sec:training}

Considering the reconstructed golden label matrix $M_{label}$ and the prediction matrix $M_{pred}$, we apply the categorical cross-entropy (CCE) loss as training loss. The CCE loss can be calculated as follows:

\begin{equation}
\small
\begin{aligned}
\mathcal{L}_{CCE}=-\frac{1}{k} \sum_{i=1}^k \sum_{c=1}^2 ( m_{label_{i}} \cdot \log \left(m_{pred_{i}}^{c}\right) + \\ \left( 1-m_{label_{i}} \right) \cdot \log \left(1-m_{pred_{i}}^{c}\right) )
\label{eq:cce_loss}
\end{aligned}
\end{equation}
where $k$ represents the number of span indexes, as well as the sequence length of the passage part;
$m_{label} \in \mathbb{B}^{k}$ is a subset of $M_{label}$, representing the label matrix corresponding to each class of entities;
$m_{pred}^{c} \in \mathbb{R}^{k \times 2}$ ($c \in \{1, 2\}$) is a subset of $M_{pred}$, representing the $c$-th value along the last dimension of the prediction matrix corresponding to each class of entities.

Then, we calculate the overall training loss by summing the CCE loss for all entity types:
\begin{equation}
\begin{aligned}
\small
\mathcal{L}_{overall}=\sum_{i=1}^{N_O} \mathcal{L}_{CCE_{i}}
\label{eq:overall_loss}
\end{aligned}
\end{equation}

\begin{algorithm}[tp]
    \renewcommand{\algorithmicrequire}{\textbf{Input:}}
    \renewcommand{\algorithmicensure}{\textbf{Output:}}
    \small
    \caption{NER-to-MRC Inference}
    \label{alg:N2M-inference}
    \begin{algorithmic}[1]
        \Require{Sequence length: $k$; Number of the options: $N_O$; Predicted matrix: $ M_{pred} \in \mathbb{R}^{k \times N_O \times 2} $; Reconstructed label matrix: $ M_{label} \in \mathbb{B}^{k \times N_O} $};
        \Ensure{Predicted entity label $p_{entity}$ with IOB-format tagging. }
        \Statex
        \State $a_{pred} \gets \argmax\limits \| M_{pred}\| $ \Comment{$a_{pred} \in \mathbb{B}^{k \times N_O}$}
        \For{$m = 1 \to k$} 
            \If{${\sum_{n=1}^{N_O} a_{pred} \left[m\right] \left[n\right]} >= 1$} \Comment{Case A}
                \State predict $\max \left(M_{pred}\left[m\right]\right)$ along the second value of the last dimension as the label on position $m$.
            \Else \Comment{Case B}
                \State predict ``O'' as the label on position $m$.
            \EndIf
        \EndFor 
        \For{$e = 1 \to k$}
            \If{${p_{entity} \left[e\right]}$ is the first entity for current entity type}
                \State Add ``B-'' before ${p_{entity} \left[e\right]}$.
            \ElsIf{${p_{entity} \left[e\right]}$ is not the first entity for current entity type}
                \State Add ``I-'' before ${p_{entity} \left[e\right]}$.
            \Else
                \State Keep ${p_{entity} \left[e\right]}$ unchanged.
            \EndIf
        \EndFor
        \State \Return{$p_{entity}$}
    \end{algorithmic}
\end{algorithm}

\subsubsection{Inference}
\label{sec:inference}

Since the output is a matrix $M_{pred} \in \mathbb{R}^{k \times N_O \times 2}$, we design a simple recovering rule to generate the entity labels, as described in~\cref{alg:N2M-inference} and~\cref{fig:finetuning}.
Recall that we remove IOB-format tagging such as``B-'' and ``I-'' from the label and convert it using one-hot encoding in~\cref{sec:reconstruction}.
Therefore, we employ a simple $argmax$ calculation on the last dimension of $M_{pred}$. After the calculation, $M_{pred} \in \mathbb{B}^{k \times N_O}$ becomes a binary matrix, where 1 indicates that the related class is selected, and 0 implies that it is not selected. The following two scenarios can be considered based on the selected conditions:

\nbf{Case A:} The case A has multiple categories with a predicted result of 1, indicating that the model considers multiple entity types as possible labels.
We choose the entity type with the highest probability as the label for the current position based on $M_{pred}$ before the $argmax$ calculation. 
When the predicted result only has one category with a value of $1$, we take the category with a result of $1$ as the entity type of the current position.

\nbf{Case B:} All categories predict $0$s. In this case, we predict the label of the current position to be ``O''.

After assigning the possible entity types for each word, we add IOB-format tagging ``B-'' and ``I-'' to the entity labels. 
Specifically, we allocate `B-'' to the first of each entity label and ``I-'' to the subsequent ones if they are consecutive identical labels.
For example, given a predicted sequence of entity types (``LOC'', ``LOC'', ``PER''), the final prediction will be ``B-LOC'', ``I-LOC'', ``B-PER''.

%% file: sections/3experiment.tex
\section{Experiments}
\label{sec:experiments}

\subsection{Experimental Setup}
\label{sec:experiments_setup}

\input{sections/table_main.tex}

\nbf{Datasets.} We evaluate our NER-to-MRC framework on several NER benchmarks that are widely used in the three domains:

\begin{itemize}
    \item \textbf{Social Media:} 
    WNUT-$16$ (WNUT $2016$ Twitter Named Entity Recognition~\cite{DBLP:conf/aclnut/StraussTRMX16/@wnut16}) dataset and WNUT-$17$ (WNUT $2017$ Emerging and Rare Entity Recognition~\cite{DBLP:conf/aclnut/DerczynskiNEL17@wnut17}) dataset are two challenging datasets in the NER field, and most SoTA methods still struggle with around $60\%$ on F1 scores. 
    Two factors make these datasets challenging, the first of which is that they contain a wide variety of entity types. 
    For example, the WNUT-$16$ dataset has $10$ types of entities, while the WNUT-$17$ dataset has $6$ types of entities. 
    In addition, they have a wide range of different entities, therefore remembering one particular entity will only be helpful for part of the task.
    \item \textbf{News:} 
    CoNLL-$2003$~\cite{DBLP:conf/conll/SangM03@conll2003} is the most commonly used NER dataset for testing the model's performance. 
    We test our model on both the CoNLL-$2003$ dataset as well as its annotation revised version, the CoNLL$++$~\cite{DBLP:conf/emnlp/WangSLLLH19@conll++} dataset.
    \item \textbf{Biomedical:}
    For the biomedical domain, we collect two typical datasets, BC$5$CDR (BioCreative V Chemical Disease Relation~\cite{DBLP:journals/biodb/LiSJSWLDMWL16@bc5cdr}) and NCBI (NCBI-disease~\cite{DBLP:journals/jbi/DoganLL14@ncbi}) dataset. 
    Moreover, we combine the train and development sets to train models by following~\citet{DBLP:conf/acl/WangJBWHHT20@clkl}.
\end{itemize}

\nbf{Evaluation metric.} 
Based on our observations, common NER tasks comprise multiple entity types, resulting in an evaluation metric across them.
Therefore, the micro-averaged F$1$ score is employed by avoiding potential unfair measurement~\cite{DBLP:journals/tkde/LiSHL22@ner-survey}.

\nbf{Main baselines.} 
Considering that our solution is a universal solution, we mainly compare a variety of baselines that are also universal models, including:

\begin{itemize}
    \item \textbf{BERT-MRC} develops a practical MRC framework with a similar idea to transform the NER task into the MRC task. 
    Unfortunately, most of the results of BERT-MRC are on proprietary datasets. Therefore the only dataset we can compare the performance of BERT-MRC with our NER-to-MRC is CoNLL-$2003$.
    \item \textbf{LUKE} is a powerful general-purpose entity representation pre-trained model. 
    It employs several entity-specific pre-training strategies and achieves SoTA performance in several entity-related domains. 
    In our experiments, we consider its sentence-level results for a fair comparison, which are reported by~\citet{DBLP:conf/acl/WangJBWHHT20@clkl}.
    \item \textbf{CL-KL} allows the model to effectively merge information retrieved by search engines and utilize several strengthened strategies, such as cooperative learning, making it one of the most competitive NER models in recent years.
\end{itemize}

\subsection{Implementation Setting}
\label{sec:implementation}

In our framework, we employ the DeBERTa-v$3$-large~\cite{DBLP:conf/iclr/HeLGC21@deberta} as backbone models in all scenarios.
Unless otherwise specified, the learning rate is $8e-6$, warming up first and then decaying linearly, and the training epoch number is $10$.
For the HRCA layers, we consider different settings of the multi-head attention to deal with different datasets.
For the challenging datasets, WNUT-$16$ and WNUT-$17$, we apply $16$ self-attention heads with $32$ dimensions in attention hidden states. 
For other datasets, we set $8$ self-attention heads with $64$ dimensions in attention hidden states. 
In our experiments, we run three times and take average scores as the final score, and they are done on a single A$100$ GPU.

\input{sections/table_ablation_backbones.tex}

\subsection{Main Results}
\label{sec:main_results}

We examine the effectiveness of our framework in the axes of i) various domains and ii) models with different designing purposes, as shown in~\cref{table:main_table}.
In summary, our NER-to-MRC framework achieves the best performance across all datasets with improvement ranging from $0.33\%$ to $11.24\%$.

Regarding domains, our framework only learns from the datasets without any extra information.
Specifically, biomedical benchmarks require expert knowledge of specific vocabulary. 
Therefore, in-domain models such as Bio-BERT address this issue by introducing biomedical corpus.
Interestingly, our general-purpose framework only employs a general-purpose PLM to solve specific problems.
Moreover, our data processing workflow does not need to change for different domains.

Considering various NER methods, the most successful ones rely on additional knowledge from pre-training data (LUKE, Bio-BERT) or the Internet (CL-KL).
In contrast, our method overcomes them with what is on hand.
There are at least two potential explanations for our improvements.
One is that our inputs include almost all information from the dataset, such as overlooked label information.
Another explanation is that the MRC reasoning strategy helps learn powerful representations and then prompts networks to generate proper choices.

BERT-MRC is a well-known model that transforms NER tasks into MRC tasks in the data reconstruction stage.
The comparison shows that our framework outperforms BERT-MRC on CoNLL-$03$.
Additionally, their input reconstruction requires hand-crafted question prompts, resulting in unstable predictions and complex extensions on other datasets.

\subsection{Ablation Study}
\label{sec:ablation_study}

\subsubsection{The effectiveness on NER-to-MRC across different backbones}
\label{sec:backbone}

As presented in~\cref{table:backbone}, we explore the mainstream PLMs as our backbone models across various domains.
Specifically, we consider DeBERTa~\cite{DBLP:conf/iclr/HeLGC21@deberta}, XLM-RoBERTa~\cite{DBLP:conf/acl/ConneauKGCWGGOZ20@xlm-roberta} in the general vocabulary (social media and news) and BioBERT~\cite{DBLP:journals/bioinformatics/LeeYKKKSK20@biobert} in the special vocabulary (biomedical). The detailed model architecture are shown in~\cref{append:backbone_architecture}.
We design the ``vanilla'' framework as the typical token-level classification tasks by the similar setting in BERT~\cite{DBLP:conf/naacl/DevlinCLT19@bert}. 
In detail, we incorporate PLMs with an additional classification layer for the hidden state corresponding to each word to generate the answer.

The results show the effectiveness of our framework, which improves all backbone models on all benchmarks.
For example, DeBERTa-v3-base earns $24.72\%$ improvement gains on WNUT-16 by introducing the NER-to-MRC framework.
In particular, the results on BC5CDR and NCBI imply that our framework also works well on domain-specific pre-training models, which verifies its cross-domain generalization capability.
Specifically, in biomedical domain, the NER-to-MRC improves general-propose DeBERTa-v3-base more than specific-propose BioBERT-base.
A possible reason is that our framework introduces the option information to assist PLMs in learning domain-specific information.

Another point worth noting is that our approach showed a relatively large improvement on the WNUT-$16$ dataset compared to other datasets, achieving an average improvement of $20.58\%$ across different backbones. WNUT-$16$ dataset has two distinctive characteristics: i) increased difficulty: compared to general datasets like CoNLL-$2003$, WNUT-$16$ includes some entity types that are very "challenging" for models, such as Sports team, TV show, Other entity, etc. These entities often have word combinations that are difficult for models to imagine, requiring higher contextual reasoning ability; ii) increased noise: the annotation guidelines for WNUT-$16$ have a lot of noise (as mentioned in~\cref{append:options}), which contaminates the labels to a certain extent. This requires models to have denoising and understanding capabilities, otherwise they cannot handle this dataset.
Our framwork has stronger reasoning and understanding capabilities, which is why it achieved significant improvement on the WNUT-$16$ dataset.

\subsubsection{Impact of the main modules}
\label{sec:mrc_vanilla}

For a comprehensive understanding of our purposed NER-to-MRC, we consider two main designed modules: 1) MRC reconstruction and 2) MRC reasoning strategy (detailed setting in~\cref{append:details_main_modules}).
\cref{table:mrc_vanilla} reports the joint effect of the aforementioned modules.
In a nutshell, both MRC reconstruction and MRC reasoning strategy cause positive effects on NER performance.
Specifically, for a challenging dataset with massive entity classes such as WNUT-$17$, MRC reasoning strategy provides more improvements than MRC reconstruction.
A possible reason is that our reasoning method includes the option content to enhance the passage tokens for final predictions.
Considering a dataset with few entity classes ($4$ types), the vanilla case has achieved a high micro-averaged F$1$ score ($94.39\%$).
Furthermore, there are two potential explanations for MRC reconstruction to help more than the MRC reasoning strategy.
One is that the PLMs can easily solve the issue with a small search space of entity labels.
Therefore, the label information is good enough for PLMs to learn. 
Another is the small promotion space to limit the potentiality of the MRC reasoning strategy.

\begin{table}[]
\begin{center}
\begin{adjustbox}{max width=0.5\textwidth}
\small
\begin{tabular}{l|c|c}
\toprule
                                         & WNUT-17         & CoNLL$++$ \\ \midrule
Vanilla                                  & 57.78           & 94.39 \\ \midrule
\;\,\,\,$+$ MRC reconstruction                   & 58.65 (+1.51\%)   & 95.01 (+0.66\%) \\ \midrule
\begin{tabular}[c]{@{}c@{}}$+$ MRC reconstruction\\ \& MRC reasoning strategy\end{tabular}                                              & \textbf{60.91 (+5.42\%)} & \textbf{95.31 +(0.97\%)} \\
\bottomrule
\end{tabular}
\end{adjustbox}
\end{center}
\caption{
Ablation over the main modules of our purposed NER-to-MRC on WNUT-$17$ and CoNLL$++$ test set. Each result is reported as a percentage of the micro-averaged F$1$ score.
}
\label{table:mrc_vanilla}
\end{table}

\begin{table}[!tp]
\begin{center}
\begin{adjustbox}{max width=0.5\textwidth}
\small
\begin{tabular}{c|c|c}
\toprule
Option content source          & WNUT-17        & CoNLL$++$ \\ \midrule
Entity names only        & 58.43          & 94.98  \\
Def. from the Int.       & 60.09          & 95.22  \\
Annotation guidelines    & \textbf{60.91} & \textbf{95.31}  \\
\bottomrule
\end{tabular}
\end{adjustbox}
\end{center}
\caption{
Results on WNUT-$17$ and CoNLL$++$ test set using different option content for the input reconstruction. Each result is reported as a percentage of the micro-averaged F$1$ score.
}
\label{table:option_type}
\end{table}

\subsubsection{Influence of option content}
\label{sec:option_type}

Introducing label information as options improves our MRC framework on NER datasets.
Therefore, \cref{table:option_type} ablates the effect of input reconstruction with different option contents.
Here, we study three option constructions with different sources: i) annotation guidelines, ii) definitions from the Internet (Def. from the Int.), and iii) Entity names only.
Specifically, annotation guidelines are the annotation descriptions of entities appended in the dataset.
For Def. from the Int., we collect the retrieval text from the Google search engine where the queries are the entity names.
The specific compositions are given in~\cref{append:option_type_ex}.

\cref{table:option_type} shows that providing the descriptions of entities prompts PLMs to learn better than just the entity names.
Providing more instructions to models will enable them to gain a deeper understanding of the text.
Moreover, the annotation guidelines work better because the annotators understand the task and construct the dataset based on this.

\begin{figure}[!tp]
\centering
\includegraphics[width=0.49\textwidth]{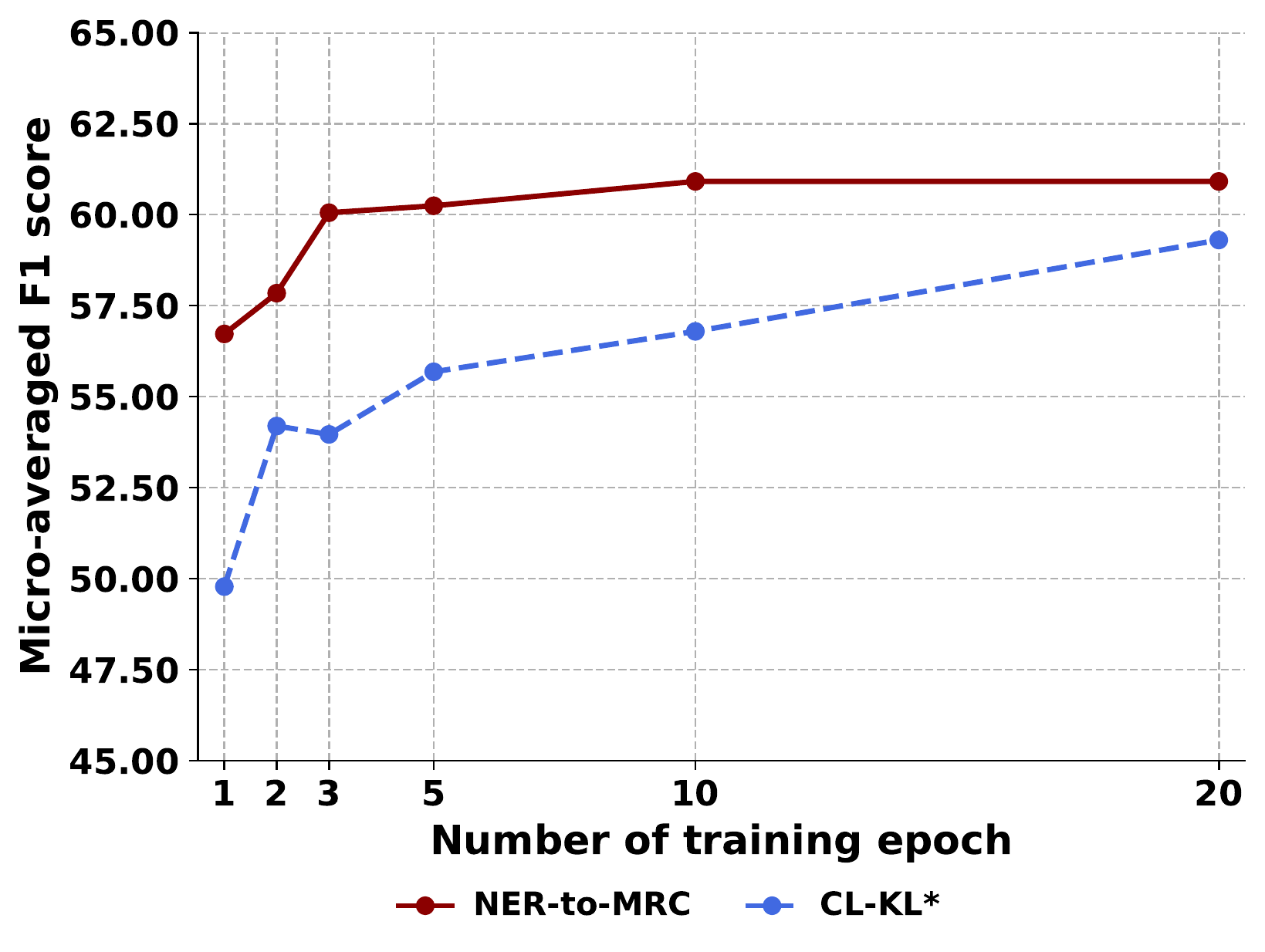}
\caption{
The performance v.s. training epochs trade-off based on different methods.
The results come from WNUT-$17$ test set.
CL-KL$^*$ indicates our reproduction of CL-KL w/o CONTEXT with open-source codes~\cite{DBLP:conf/acl/WangJBWHHT20@clkl}.
For a stable result, we take the average score of $3$ runs.
Under a fair environment, we align the batch size as $2$ for them.
}
\label{fig:convergence}
\end{figure} 

\subsubsection{Convergence speed}
\label{sec:convergence}

We explore the detailed comparisons of the performance v.s. training epochs trade-off as shown in~\cref{fig:convergence}.
Our framework only passes $3$ epochs to converge around $60.00$ percent of the micro-averaged F1 score.
In contrast, CL-KL$^*$ requires at least $20$ epochs to converge and presents an unstable disturbance on early $5$ epochs.
The outcome verifies that our method learns better while viewing fewer samples, which implies better usability for further deployment.

%% file: sections/table_main.tex
\begin{table*}[!tp]
\begin{center}
\begin{adjustbox}{max width=\textwidth}
\small
\begin{tabular}{l|cc|cc|cc}
\toprule
                   & \multicolumn{2}{c|}{\textbf{Social Media}}       & \multicolumn{2}{c|}{\textbf{News}}               & \multicolumn{2}{c}{\textbf{Biomedical}}                \\ 
\midrule
                  & WNUT-16        & WNUT-17                & CoNLL-03       & CoNLL++                & BC5CDR                & NCBI                  \\ 
\midrule
{\ul In-domain models}  & & & & & & \\
DATNet-F~\cite{DBLP:conf/acl/ZhouZJZFGK19@datnet}                  & 53.43          & 42.83                  & -              & -                      & -                     & -                     \\
InferNER~\cite{DBLP:conf/flairs/ShahzadAEN21@inferner}                 & 53.48          & 50.52                  & 93.76          & -                      & -                     & -                     \\
BERTweet~\cite{DBLP:conf/emnlp/NguyenVN20@bertweet}                  & 52.10          & 56.50                  & -              & -                      & -                     & -                     \\
SA-NER~\cite{DBLP:conf/emnlp/NieTWSD20@saner}                   & 55.01          & 50.36                  & -              & -                      & -                     & -                     \\
RoBERTa-large + RegLER~\cite{jeong2021regularizing@regler}    & -              & 58.30                  & 92.30          & -                      &                       &                       \\
CrossWeigh + Pooled Flair~\cite{DBLP:conf/emnlp/WangSLLLH19@crossweight} & -              & 50.03                  & 93.43          & 94.28                  & -                     & -                     \\
ACE-sentencce-level~\cite{DBLP:conf/acl/WangJBWHHT20a@ace}                 & -              & -                      & {\ul 93.60}          & -                      & -                     & -                     \\
NER+PA+RL~\cite{DBLP:conf/acl-deeplo/NooralahzadehLO19@nerparl}                 & -              & -                      & -              & -                      & 89.93                 & -                     \\
Bio-Flair~\cite{DBLP:journals/corr/abs-1908-05760@bioflair}                 & -              & -                      & -              & -                      & 89.42                 & 88.85                 \\
Bio-BERT~\cite{DBLP:journals/bioinformatics/LeeYKKKSK20@biobert}                  & -              & -                      & -              & -                      & -                     & 87.70                 \\
Bio-BERT + RegLER~\cite{jeong2021regularizing@regler}          & -              & -                      & -              & -                      & -                     & 89.00                 \\ 
Bio-BERT + KGQA~\cite{DBLP:journals/health/BanerjeePDB21@KGQA}                 & -              & -                      & -          & -                      & 89.21                     & 89.05                     \\ \midrule
{\ul Universal models}  & & & & & & \\
LUKE~\cite{DBLP:conf/emnlp/YamadaASTM20@luke}                      & 54.04          & 55.22                  & 92.42          & 93.99                  & 89.18                 & 87.62                 \\
BERT-MRC~\cite{DBLP:conf/acl/LiFMHWL20@bertmrc,DBLP:journals/health/BanerjeePDB21@KGQA}                 & -              & -                      & 93.04          & -                      & 72.43                     & 75.19                     \\
CL-KL w/o CONTEXT~\cite{DBLP:conf/acl/WangJBWHHT20@clkl}        & 58.14          & 59.33                  & 93.21          & 94.55                  & 90.73                 & {\ul 89.24}                 \\
CL-KL w CONTEXT~\cite{DBLP:conf/acl/WangJBWHHT20@clkl}       & {\ul 58.98}          & {\ul 60.45}                 &  93.56          & {\ul 94.81}                  & {\ul 90.93}                 & 88.96                 \\
\midrule
NER-to-MRC (ours) & \textbf{65.61} & \textbf{60.91} & \textbf{93.91} & \textbf{95.23} & \textbf{92.25} & \textbf{90.38}                 \\
\bottomrule
\end{tabular}
\end{adjustbox}
\end{center}
\caption{
Comparisons of our proposed model with previous SoTA results on various NER datasets. Each result is reported as a percentage of the micro-average F1 score. The best scores are in \textbf{bold}, and the second best scores are {\ul underlined}.
}
\label{table:main_table}
\end{table*}

%% file: sections/table_ablation_backbones.tex
\begin{table*}[tp]
\begin{center}
\begin{adjustbox}{max width=\textwidth}
\small
\begin{tabular}{lc|cc|cc|cc}
\toprule
                                                       &            & \multicolumn{2}{c|}{\textbf{Social Media}} & \multicolumn{2}{c|}{\textbf{News}}        & \multicolumn{2}{c}{\textbf{Biomedical}} \\ \midrule
                                                       &            & WNUT-16              & WNUT-17             & CoNLL-03       & CoNLL++                  & BC5CDR             & NCBI               \\ \midrule
\multicolumn{1}{l|}{\multirow{2}{*}{\begin{tabular}[c]{@{}l@{}}DeBERTa-v3-base\\ \cite{DBLP:conf/iclr/HeLGC21@deberta}\end{tabular}}}  & Vanilla    & 52.10                & 55.93               & 91.43          & 94.11                    & 88.99                  & 85.01                   \\
\multicolumn{1}{l|}{}                                  & NER-to-MRC & 64.98 (+24.72\%)     & 58.85 (+5.22\%)     & 93.57 (+2.34\%)& 94.92 (+0.86\%)          & 91.53 (+2.85\%)                  & 87.69 (+3.15\%)                  \\ \midrule
\multicolumn{1}{l|}{\multirow{2}{*}{\begin{tabular}[c]{@{}l@{}}XLM-RoBERTa-large\\ \cite{DBLP:conf/acl/ConneauKGCWGGOZ20@xlm-roberta}\end{tabular}}} & Vanilla    & 53.32                & 56.74               & 92.31          & 93.39                    & -                  & -                  \\
\multicolumn{1}{l|}{}                                  & NER-to-MRC & 64.31 (+15.96\%)     & 58.01 (+2.24\%)     & 93.39 (+1.17\%)& 94.54 (+1.23\%)          & -                  & -                  \\ \midrule
\multicolumn{1}{l|}{\multirow{2}{*}{\begin{tabular}[c]{@{}l@{}}BioBERT-base\\ \cite{DBLP:journals/bioinformatics/LeeYKKKSK20@biobert}\end{tabular}}}     & Vanilla    & -                    & -                   & -              & -                        & 89.44              & 86.16             \\
\multicolumn{1}{l|}{}                                  & NER-to-MRC & -                    & -                   & -              & -                        & 90.85 (+1.58\%)    & 88.01 (+2.15\%)    \\ \midrule
\multicolumn{1}{l|}{\multirow{2}{*}{\begin{tabular}[c]{@{}l@{}}DeBERTa-v3-large\\ \cite{DBLP:conf/iclr/HeLGC21@deberta}\end{tabular}}} & Vanilla    & 54.19   & 57.78   & 92.57   & 94.39  & 90.76  & 88.83  \\
\multicolumn{1}{l|}{}                                  & NER-to-MRC & \textbf{65.61 (+21.07\%)} & \textbf{60.91 (+5.42\%)} & \textbf{93.91 (+1.45\%)} & \textbf{95.31 (+0.97\%)} & \textbf{92.25 (+1.64\%)} & \textbf{90.38 (+1.74\%)}     \\ \bottomrule
\end{tabular}
\end{adjustbox}
\end{center}
\caption{
The results by deploying NER-to-MRC framework across different backbones. Each result is reported as a percentage of the micro-averaged F$1$ score. The best performance for each dataset is shown in \textbf{bold}.
}
\label{table:backbone}
\end{table*}

%% file: sections/4conclusion.tex
\section{Conclusions}
\label{sec:conclusion}

In this paper, we have proposed a complete NER-to-MRC conversion by considering NER problems via a MRC perspective.
Our approach first reconstructs the NER data into MRC inputs, and then applies the MRC reasoning strategy to predict a rational choice.
Furthermore, it shows the state-of-the-art performance on $6$ benchmarks across three domains.
Note that this success is based on a single general-purpose PLM without external data.
Moreover, our experimental results demonstrate that the NER-to-MRC framework is compatible with a set of different PLMs and that our design is efficient in terms of performance enhancement and convergence speed.

%% file: sections/5limitations.tex
\section*{Limitations}
\label{sec:limitations}

Most existing NER datasets do not feature nested tags~\cite{DBLP:conf/coling/YadavB18@flatnested}. In a nested NER dataset, multiple entities can be found inside an entity.
Though we provide a NER-to-MRC framework with generalization ability in the NER task, we only evaluate our method with flat NER datasets, which might lead to a less comprehensive scope of our benchmarks.
Our solution can be naturally and conveniently applied to handle nested NER tasks. Specifically, we only need to set a threshold so that each word outputs a label for each entity category, allowing a word to have multiple labels to handle nested NER problems.
In the future, we will provide the related instructions and further evaluate more NER datasets.

%% file: sections/6ethical.tex
\section*{Ethical Considerations}
\label{sec:ethical}

Our NER-to-MRC framework brings a powerful tool to the real-world across multiple domains such as social media, news, and biomedical.
Therefore, the ethical influence of this work might spread to many applications.
The ethical implications involve two main points: i) the bias from the backbone networks and ii) training datasets.
For the networks, several analyses point out that ethnic biases are included in the PLMs such as BERT~\cite{DBLP:journals/corr/abs-2211-14402@biases-bert} and GPT3~\cite{lucy-bamman-2021-gender@baiases-gpt3}.
The potential risks are unpredictable after deployments with those PLMs.
Fortunately, our backbones are replaceable as described in~\cref{sec:network}.
Therefore, we encourage the users to install unbiased language models and provide model cards for the details.
Beyond the backbone PLMs, it is necessary to pay attention to our downstream tasks, such as gender, race, and sexual orientation.
In real-world deployments, we suggest it is necessary to design a slew of cleaning procedures such as SampleClean and CPClean~\cite{DBLP:journals/corr/abs-2109-07127@cleaning-methods}.
After that, we encourage open discussions about its utilization, hoping to reduce potential malicious behaviors.

%% file: sections/appendix.tex
\appendix

\section{Dataset Details}
\label{append:dataset_details}

\subsection{Dataset statistics}
\label{append:dataset_statistics}

We summarize the statistics of the datasets used in this paper in~\cref{table:dataset_statistics}. 
Specifically, ``Avg. Length'' implies the average of the lengths across all dataset splits with including train set, development set, and test set.
Since our method create options with label information as described in~\cref{sec:reconstruction}, we collect their average lengths.
\subsection{Examples of input reconstruction}
\label{append:example_input_rec}

As presented in~\cref{table:input_ex}, we outline a sample reconstructed MRC triplet of passage, question, and options from the WNUT-$17$ dataset.

\section{Experimental Settings}

\subsection{Evaluation metric library}
\label{append:metirc_library}

In~\cref{sec:experiments_setup}, we explained how we selected the evaluation metric. To enable easy and comprehensive comparison of both past and future schemes, we computed all of the span-based micro F1 score results presented in this paper using seqeval\footnotemark[1], an open-source Python framework for sequence labeling evaluation.

\subsection{Backbone model architecture}
\label{append:backbone_architecture}

We test the performance of our NER-to-MRC framework using different PLMs as the backbone model in~\cref{sec:backbone}, including DeBERTa-V3-base, DeBERTa-V3-large, XLM-RoBERTa-large and BioBERT-base. 
We give the detailed model architectures as follows:

\nbf{DeBERTa-V3-base}: Number of Layers $= 12$, Hidden size $= 768$, Attention heads $= 12$, Total Parameters = $86$M.

\nbf{DeBERTa-V3-large}: Number of Layers $= 24$, Hidden size $= 1024$, Attention heads $= 12$, Total Parameters = $304$M.

\nbf{XLM-RoBERTa-large}: Number of Layers $= 24$, Hidden size $= 1024$, Attention heads $= 16$, Total Parameters = $355$M.

\nbf{BioBERT-base}: Number of Layers $= 12$, Hidden size $= 768$, Attention heads $= 12$, Total Parameters = $110$M.

\subsection{Ablation experiment setup}
\label{append:details_main_modules}

In~\cref{sec:mrc_vanilla}, we perform ablation experiments for the main modules of our proposed NER-to-MRC. 
The detailed settings are:

\nbf{Vanilla}: we fine-tune the hidden state of the PLMs.

\nbf{$+$ MRC reconstruction}: we reconstruct the NER dataset into the MRC reconstruction. However, this setting plugs a MLP layer to do multiple choice without MRC reasoning strategy.

\nbf{$+$ MRC reconstruction \& MRC reasoning strategy}: it follows the same instructions as the NER-to-MRC desgin.

\subsection{Sources in our inputs}
\label{append:options}

As discussion in~\cref{sec:reconstruction}, we take the same descriptions in dataset papers or their released homepages.
We apply this setting on WNUT-$17$, CoNLL-$2003$, CoNLL$++$, BC$5$CDR and NCBI. For all the datasets we used in this paper, we follow the license terms of the corresponding papers.
In particular, WNUT-$16$ dataset only provide a google document\footnotemark[2] as the annotation guidelines.
Unfortunately, this document is inadequate in entity types and contains a lot of noisy text, resulting not applicable label information.
Therefore, we collect the definitions of each entity from the Internet, which is the same approaches in~\cref{sec:option_type}.
In detail, we put the full information in~\cref{table:wnut16_option}.

\subsection{Details of different option types}
\label{append:option_type_ex}

In~\cref{sec:option_type}, we ablate the impact of different sources of option type composition on the model performance. The annotation guidelines can be easily found and utilized in the paper or homepage corresponding to the dataset. Consequently, we do not repeat the examples of annotation guidelines compositions. Instead, we only demonstrate examples of compositions defined from the Internet and examples of compositions using only entity names for the WNUT-$17$ dataset in~\cref{table:option_type_ex_wnut17} and the CoNLL$++$ dataset in~\cref{table:option_type_ex_conll}.

\footnotetext[1]{https://github.com/chakki-works/seqeval}
\footnotetext[2]{https://docs.google.com/document/d/12hI-2A3vATMWRdsKkzDPHu5oT74\_tG0-PPQ7VN0IRaw}

\section{Inference speed}
\label{append:inference_speed}

In order to simultaneously address the flat NER and nested NER tasks, our proposed approach predicts the probability of each entity type separately. Considering that inference time is an important consideration in NER task, our strategy may exhibit less-than-optimal performance in terms of inference speed compared to those approaches that handle all entity types together. Therefore, we compared several baseline approaches, including traditional BiLSTM~\cite{DBLP:conf/naacl/LampleBSKD16@bilstm_crf}, PLM + BiLSTM scheme~\cite{DBLP:conf/bmei/DaiWNLLB19@BERT-BiLSTM}, and the current state-of-the-art approach CL-KL, in terms of their inference speed and accuracy on the WNUT-17 dataset in~\cref{table:inference_speed}.

\begin{table}[]
\begin{center}
\begin{adjustbox}{max width=0.5\textwidth}
\small
\begin{tabular}{l|c|c}
\toprule
Model          & Inference speed (samples/s)  & Span-based F1 \\ \midrule
Single BiLSTM  & 64                           & 43.10 \\ \midrule
{\begin{tabular}[c]{@{}c@{}}XLM-RoBERTa-large \\ + BiLSTM \end{tabular}}   & 23   & 57.8 \\ \midrule
NER-to-MRC      & 22 & 60.91 \\ \midrule
CL-KL           & 18 & 60.45 \\ \midrule
\bottomrule
\end{tabular}
\end{adjustbox}
\end{center}
\caption{
Inference speed and performance comparison on WNUT-$17$ test set. Each F$1$ result is reported as a percentage of the micro-averaged F$1$ score.
}
\label{table:inference_speed}
\end{table}

We conducted tests on all of the aforementioned results using our own implementation. Based on the results, our approach has a similar inference speed to XLM-RoBERTa-large+BiLSTM, and the model's inference speed is faster than CL-KL. It is worth mentioning that compared to CL-KL, our training cost (our: 3 epoch v.s CL-KL 20 epoch) is relatively better. Our method achieves a better computational resources v.s. performance trade-off.

\begin{table*}[]
\begin{center}
\small
\begin{tabular}{l | l | c c c c c c}
\toprule
\multicolumn{1}{c|}{Domain}   & \multicolumn{1}{c|}{Dataset} & \# Train set & \# Dev set & \# Test set & \# Entity Types & Avg. Length & {\begin{tabular}[c]{@{}c@{}}Avg. Options \\ Length\end{tabular}} \\ \midrule
\multirow{2}{*}{Social Media} & WNUT-16                      & 2,394         & 1,000       & 3,850        & 10              & 17.2                                                           & 19.0                                     \\
                              & WNUT-17                      & 3,394         & 1,009       & 1,287        & 6               & 17.9                                                           & 33.8                                     \\ \midrule
\multirow{2}{*}{News}         & CoNLL-2003                   & 14,041        & 3,250       & 3,453        & 4               & 14.5                                                           & 156.5                                     \\
                              & CoNLL++                      & 14,041        & 3,250       & 3,453        & 4               & 14.5                                                           & 156.5                                     \\ \midrule
\multirow{2}{*}{Biomedical}   & BC5CDR                       & 5,228         & 5,330       & 5,865        & 2               & 20.4                                                           & 367.0                                     \\
                              & NCBI                         & 5,432         & 923        & 940         & 1               & 25.3                                                            & 550.0                                     \\ \bottomrule
\end{tabular}
\end{center}
\caption{
The statistics of the datasets.
}
\label{table:dataset_statistics}
\end{table*}

\begin{table*}[]
\begin{center}
\small
\begin{tabular}{p{1.9\columnwidth}}
\toprule
\textbf{Passage:} \\
\midrule
Watched your video ! Great work ! Thumbs up ! ! ! Welcome to my channel ! \\
\midrule
\textbf{Question:} \\
\midrule
What kind of entity is this? \\
\midrule
\textbf{Options:} \\
\midrule
- Names of corporations (e.g. Google). Don’t mark locations that don’t have their own name. Include punctuation in the middle of names. \\
- Names of creative works (e.g. Bohemian Rhapsody). Include punctuation in the middle of names. The work should be created by a human, and referred to by its specific name. \\
- Names of groups (e.g. Nirvana, San Diego Padres). Don’t mark groups that don’t have a specific, unique name, or companies (which should be marked corporation). \\
- Names that are locations (e.g. France). Don’t mark locations that don’t have their own name. Include punctuation in the middle of names. Fictional locations can be included, as long as they’re referred to by name (e.g. Hogwarts). \\
- Names of people (e.g. Virginia Wade). Don’t mark people that don’t have their own name. Include punctuation in the middle of names. Fictional people can be included, as long as they’re referred to by name (e.g. Harry Potter). \\
- Name of products (e.g. iPhone). Don’t mark products that don’t have their own name. Include punctuation in the middle of names. Fictional products can be included, as long as they’re referred to by name (e.g. Everlasting Gobstopper). It’s got to be something you can touch, and it’s got to be the official name. \\
\bottomrule
\end{tabular}
\end{center}
\caption{
An example of the inputs on WNUT-$17$ dataset.
}
\label{table:input_ex}
\end{table*}

\begin{table*}[]
\begin{center}
\small
\begin{tabular}{p{1.9\columnwidth}}
\toprule
\textbf{Options composition for the WNUT-$16$ dataset} \\
\midrule
- Compnay entities, or business entities, describes any organization formed to conduct business. \\
- Facility, or facilities are places, buildings, or equipments used for a particular purpose or activity. \\
- Geolocation refers to the use of location technologies such as GPS or IP addresses to identify and track the whereabouts of connected electronic devices. \\
- Musicartist is One who composes, conducts, or performs music, especially instrumental music. \\
- Other entities are entities other than company, facility, geolocation, music artist, person, product, sports team and tv show. \\
- Person entities are named persons or family. \\
- Product entities are name of products (e.g. iPhone**) which you can touch it, buy it and it's the technical or manufacturer name for it. Not inclduing products that don't have their own name. Include punctuation in the middle of names.,
- Sports team is a group of individuals who play sports (sports player). \\
- Tv show is any content produced for viewing on a television set which can be broadcast via over-the-air, satellite, or cable, excluding breaking news, advertisements, or trailers that are typically placed between shows. \\
\bottomrule
\end{tabular}
\end{center}
\caption{
Specific composition of the options part on WNUT-$16$ dataset.
}
\label{table:wnut16_option}
\end{table*}

\begin{table*}[]
\begin{center}
\small
\begin{tabular}{p{1.9\columnwidth}}
\toprule
\textbf{Option type: Def. from the Int.} \\
\midrule
- Corporate entities are business structures formed specifically to perform activities, such as running an enterprise or holding assets. Although it may be comprised of individual directors, officers, and shareholders, a corporation is a legal entity in and of itself. \\
- Creative work entities are performance, musical composition, exhibition, writing (poetry, fiction, script or other written literary forms), design, film, video, multimedia or other new media technologies and modes of presentation. \\
- Group entities are specific, unique names, or companies. \\
- Location entities are the name of politically or geographically defined locations such as cities, provinces, countries, international regions, bodies of water, mountains, etc. \\
- Person entities are named persons or family. \\
- Product entities are name of products (e.g. iPhone**) which you can touch it, buy it and it's the technical or manufacturer name for it. Not inclduing products that don't have their own name. Include punctuation in the middle of names. \\
\midrule
\textbf{Option type: Entity name only} \\
\midrule
- Corporate\\
- Creative-work\\
- Group\\
- Location\\
- Person\\
- Product\\
\bottomrule
\end{tabular}
\end{center}
\caption{
Specific composition of the three different option types on WNUT-$17$ dataset.
}
\label{table:option_type_ex_wnut17}
\end{table*}

\begin{table*}[]
\begin{center}
\small
\begin{tabular}{p{1.9\columnwidth}}
\toprule
\textbf{Option type: Def. from the Int.} \\
\midrule
- Person entities are named persons or family. \\
- Organization entities are limited to named corporate, governmental, or other organizational entities. \\
- Location entities are the name of politically or geographically defined locations such as cities, provinces, countries, international regions, bodies of water, mountains, etc. \\
- Miscellaneous entities include events, nationalities, products and works of art. \\
\midrule
\textbf{Option type: Entity name only} \\
\midrule
- Person \\
- Organization \\
- Location \\
- Miscellaneous\\
\bottomrule
\end{tabular}
\end{center}
\caption{
Specific composition of the three different option types on CoNLL$++$ dataset.
}
\label{table:option_type_ex_conll}
\end{table*}